\documentclass{article}

\usepackage{graphicx}
\usepackage{cite}
\usepackage{apacite}

\usepackage{color}
\usepackage{amsmath}
\usepackage{amssymb}
\usepackage{amsfonts}
\usepackage{epsfig}
\usepackage{url}
\usepackage{bm}

\newtheorem{thm}{Theorem}
\newtheorem{lem}{Lemma}

\newcommand{\qed}{\fbox{\rule[1pt]{0pt}{0pt}}}
\newcommand{\bsquare}{\hbox{\rule{6pt}{6pt}}}

\def\bfa{\mbox{\boldmath $a$}}
\def\bfe{\mbox{\boldmath $e$}}
\def\bff{\mbox{\boldmath $f$}}
\def\bfx{\mbox{\boldmath $x$}}
\def\bfr{\mbox{\boldmath $r$}}
\def\bfw{\mbox{\boldmath $w$}}

\def\bsigma{\mbox{\boldmath $\sigma$}}

\newcommand{\A}{{\bf A}}
\newcommand{\B}{{\bf B}}
\newcommand{\C}{{\bf C}}
\newcommand{\I}{{\bf I}}
\newcommand{\D}{{\bf D}}
\newcommand{\x}{{\bf x}}
\newcommand{\X}{{\bf X}}
\newcommand{\R}{{\bf R}}

\newcommand{\bLambda}{{\bf \Lambda}}

\title{ParceLiNGAM: A causal ordering method robust against latent confounders}
\author{Tatsuya Tashiro\thanks{The Institute of Scientific and Industrial Research (ISIR), Osaka University,
Mihogaoka 8-1, Ibaraki, Osaka 567-0047, Japan. Email: tashiro@ar.sanken.osaka-u.ac.jp}, Shohei Shimizu\thanks{Osaka University, Japan}, Aapo Hyv\"arinen\thanks{University of Helsinki, Finland} and
Takashi Washio\thanks{Osaka University, Japan}}
\date{}

\begin{document}

\maketitle

\begin{center} {\bf Abstract} \end{center}
We consider learning a causal ordering of variables in a linear non-Gaussian acyclic model called LiNGAM. 
Several existing methods have been shown to consistently estimate a causal ordering assuming that all the model assumptions are correct. 
But, the estimation results could be distorted if some assumptions actually are violated. 
In this paper, we propose a new algorithm for learning causal orders that is robust against one typical violation of the model assumptions: latent confounders.
The key idea is to detect latent confounders by testing independence between estimated external influences and find subsets (parcels) that include variables that are not affected by latent confounders.
We demonstrate the effectiveness of our method using artificial data and simulated brain imaging data.

\section{Introduction}\label{sec:intro}
Bayesian networks have been widely used to analyze causal relations of variables in many empirical sciences \cite{Bollen89book,Pearl00book,Spirtes93book}. 
A common assumption is linear-Gaussianity. 
But this poses serious identifiability problems so that many important models are indistinguishable with no prior knowledge on the structures. 
Recently, it was shown by \cite{Shimizu06JMLR} that the utilization of non-Gaussianity allows the full structure of a linear acyclic model to be identified without pre-specifying any causal orders of variables. 
The new model, a Linear Non-Gaussian Acyclic Model called LiNGAM \cite{Shimizu06JMLR}, is closely related to independent component analysis (ICA) \cite{Hyva01book}.

Most existing estimation methods \cite{Shimizu06JMLR,Shimizu11JMLR,Hyva13JMLR} for LiNGAM learn causal orders assuming that all the model assumptions hold. 
Therefore, these algorithms could return completely wrong estimation results when some of the model assumptions are violated. 
Thus, in this paper, we propose a new algorithm for learning causal orders that is robust against one typical model violation, {\it i.e.}, latent confounders.  
A latent confounder means a variable which is not observed but which exerts a causal influence on some of the observed variables. 
Many real-world applications including brain imaging data analysis \cite{Smith11NI} could benefit from our approach. 

This paper\footnote{Some preliminary results were presented in \cite{Tashiro12ICANN}, which corresponds to Section~\ref{sec:est} of this paper.} is organized as follows. 
We first review LiNGAM \cite{Shimizu06JMLR} and its extension to latent confounder cases \cite{Hoyer07IJAR} in Section~\ref{sec:lvlingam}. 
In Section~\ref{sec:proposal}, we propose a new algorithm to learn causal orders in LiNGAM with latent confounders. 
We empirically evaluate the performance of our algorithm using artificial data in Section~\ref{sec:exp} and simulated fMRI data in Section~\ref{sec:real}.
We conclude this paper in Section~\ref{sec:conc}. 

\section{Background: LiNGAM with latent confounders}\label{sec:lvlingam}
We briefly review a linear non-Gaussian acyclic model called LiNGAM \cite{Shimizu06JMLR} and an extension of the LiNGAM to cases with latent confounding variables \cite{Hoyer07IJAR}. 

In LiNGAM \cite{Shimizu06JMLR}, causal relations of observed variables $x_i$ ($i=1,\cdots,d$) are modeled as: 
\begin{eqnarray}
x_i &=& \sum_{k(j)<k(i)} b_{ij}x_j + e_i, \label{eq:model1}
\end{eqnarray}
where $k(i)$ is a causal ordering of the variables $x_i$. 
In this ordering, the variables $x_i$ graphically form a directed acyclic graph (DAG) so that no later variable determines, {\it i.e.}, has a directed path on any earlier variable. 
The $e_i$ are external influences, and $b_{ij}$ are connection strengths. 
In matrix form, the model (\ref{eq:model1}) is written as
\begin{eqnarray}
\bfx &=& \B\bfx + \bfe,\label{eq:model2}
\end{eqnarray}
where the connection strength matrix $\B$ collects $b_{ij}$ and the vectors $\bfx$ and $\bfe$ collect $x_i$ and $e_i$.  
Note that the matrix $\B$ can be permuted  to be lower triangular with all zeros on the diagonal if simultaneous equal row and column permutations are made according to a causal ordering $k(i)$ because of the acyclicity.
The zero/non-zero pattern of $b_{ij}$ corresponds to the absence/existence pattern of directed edges. 
External influences $e_i$ follow non-Gaussian continuous distributions with zero mean and non-zero variance and are mutually independent. 
The non-Gaussianity assumption on $e_i$ enables identification of a causal ordering $k(i)$ 
based on data $\bfx$ only \cite{Shimizu06JMLR}. 
This feature is a major advantage over conventional Bayesian networks based on the Gaussianity assumption on $e_i$ \cite{Spirtes93book}. 

Next, LiNGAM with latent confounders \cite{Hoyer07IJAR} can be formulated as follows:
\begin{eqnarray}
\bfx &=& \B\bfx + \bLambda \bff + \bfe,\label{eq:lvlingam}
\end{eqnarray}
where the difference with LiNGAM in Eq.~(\ref{eq:model2}) is the existence of latent confounding variable vector $\bff$. 
A latent confounding variable is a latent  variable that is a parent of more than one observed variable. 
The vector $\bff$ collects non-Gaussian latent confounders $f_j$ with zero mean and non-zero variance $(j=1, \cdots, q)$. Without loss of generality \cite{Hoyer07IJAR}, latent confounders $f_j$ are assumed to be mutually independent. 
The matrix $\bLambda$ collects $\lambda_{ij}$ which denotes the connection strength from $f_{j}$ to $x_{i}$. 
For each $j$, at least two $\lambda_{ij}$ are non-zero 
since a latent confounder is defined to have at least two children \cite{Hoyer07IJAR}. 
The matrix $\bLambda$ is assumed to be of full column rank. 

The central problem of causal discovery based on the latent variable LiNGAM in Eq.~(\ref{eq:lvlingam}) is to estimate {\it as many} of causal orders $k(i)$ and connection strengths $b_{ij}$ {\it as possible} based on data $\bfx$ only. 
This is because in many cases only an equivalence class of the true model whose members produce the exact same observed distribution is identifiable \cite{Hoyer07IJAR}. 

In \cite{Hoyer07IJAR}, an estimation method based on overcomplete ICA \cite{Lewicki00NECO} was proposed. 
However, overcomplete ICA methods are often not very reliable and get stuck in local optima. 
Thus, in \cite{Entner10AMBN}, a method that does not use overcomplete ICA was proposed to first find variable {\it pairs} that are not affected by latent confounders and then estimate a causal ordering of one to the other. However, their method does not estimate a causal ordering of more than two variables. 
A simple cumulant-based method for estimating the model in the case of Gaussian latent confounders was further proposed by \cite{Chen13NECO}.

\section{A method robust against latent confounders}\label{sec:proposal}
In this section, we propose a new approach for estimating causal orders of more than two variables without explicitly modeling latent confounders. 

\subsection{Identification of causal orders of variables that are {\itshape not} affected by latent confounders}\label{sec:est}
We first provide principles to identify an exogenous (root) variable and a sink variable  which are such that are not affected by latent confounders in the latent variable LiNGAM in Eq.~(\ref{eq:lvlingam}) (if such variables exist) and next present an estimation algorithm. 
Recent estimation methods \cite{Shimizu11JMLR} for LiNGAM in Eq.~(\ref{eq:model2}) and its nonlinear extension \cite{Hoyer09NIPS,Mooij09ICML} learn a causal ordering by finding causal orders one by one either from the top downward or from the bottom upward assuming no latent confounders. 
We extend these ideas to latent confounder cases. 

We first generalize Lemma~1 of \cite{Shimizu11JMLR} for the case of latent confounders. 
\begin{lem}\label{lemma1}
Assume that all the model assumptions of the latent variable LiNGAM in Eq.~ (\ref{eq:lvlingam}) are met and the sample size is infinite. 
Denote by $r_i^{(j)}$ the residuals  when $x_i$ are regressed on $x_j$: 
$r_i^{(j)} = x_i - \frac{{\rm cov}(x_i,x_j)}{{\rm var}(x_j)}x_j$ $(i \neq j)$.
Then a variable $x_j$ is an exogenous variable in the sense that it has no parent observed variable nor latent confounder if and only if $x_j$ is independent of its residuals $r_i^{(j)}$ for all $i \neq j$. \mbox{\hfill \qed}
\end{lem} 
Next, we generalize the idea of \cite{Mooij09ICML} for the case of latent confounders. 
\begin{lem}\label{lemma2}
Assume that all the model assumptions of the latent variable LiNGAM in Eq.~ (\ref{eq:lvlingam}) are met and the sample size is infinite. Denote by $\bfx_{(-j)}$ a vector that contains all the variables other than $x_j$.  
Denote by $r_j^{(-j)}$ the residual  when $x_j$ is regressed on $\bfx_{(-j)}$, {\it i.e.}, 
$r_{j}^{(-j)} = x_j - \bsigma^T_{(-j)j} \Sigma_{(-j)}^{-1} \bfx_{(-j)}, $
where 
$\Sigma = 
\left[
\begin{array}{cc}
\sigma_{j} & \bsigma_{j(-j)}^T \\
\bsigma_{j(-j)} & \Sigma_{(-j)}
\end{array}
\right]$
is the covariance matrix of $[x_j, \bfx_{(-j)}^T]^T$. 
Then a variable $x_j$ is a sink variable in the sense that it has no child observed variable nor latent confounder if and only if $\bfx_{(-j)}$ is independent of its residual $r_j^{(-j)}$. \mbox{\hfill \qed}
\end{lem} 
The proofs of these lemmas are given in the appendix.\footnote{We prove the lemmas without assuming the faithfulness \cite{Spirtes93book} unlike our previous work \cite{Tashiro12ICANN}.} 

Thus, we can take a hybrid estimation approach that uses these two principles.  
We first identify an exogenous variable by finding a variable that is most independent of its residuals and remove the effect of the exogenous variable from the other variables by regressing it out. 
We repeat this until independence between every variable and all of its residuals is statistically rejected. 
Dependency between every variable and any of its residuals implies that an exogenous variable as defined in Lemma~\ref{lemma1} does not exist or some model assumption of latent variable LiNGAM in Eq.~(\ref{eq:lvlingam}) is violated. 
Similarly, we next identify a sink variable in the remaining variables by finding a variable such that its regressors and its residual are most independent and disregard the sink variable. 
We repeat this until independence is statistically rejected for every variable.\footnote{The issue of multiple comparisons arises in this context, which we would like to study in future work.}  
To test independence, we first evaluate pairwise independence between variables and the residuals using a kernel-based independence measure called HSIC \cite{GreFukTeoSonSchSmo08} and then combine the resulting $p$-values $p_i$ ($i=1,\cdots, c$) using a well-known Fisher's method \cite{Fisher50book} to compute the test statistic $-2\sum_{i=1}^c\log p_i$, which follows the chi-square distribution with $2c$ degrees of freedom when all the pairs are independent. 

Since all the causal orders are not necessarily identifiable in the latent variable LiNGAM in Eq.~(\ref{eq:lvlingam}) \cite{Hoyer07IJAR},
 we here aim to estimate a $d \times d$ causal ordering matrix $\C$$=$$[c_{ij}]$ that collects causal orderings between two variables, which is defined as
\begin{eqnarray}
c_{ij}&:=&\left\{ 
\begin{array}{rl}
-1 & \ {\rm if} \ k(i) < k(j) \\
1 & \ {\rm if} \ k(i) > k(j)\\
0 & \ {\rm if \ it\ is\ unknown\ whether\ either\ of \ the\ two\ cases}\\
 & \ {\rm  \ above\ (-1\ or\ 1)\ is\ true}.
\end{array}
\right.
\end{eqnarray}
Thus, the estimation consists of the following steps: 
\noindent \\
  \rule{\columnwidth}{0.5mm}
	Algorithm~1:  Hybrid estimation of causal orders of variables that are not affected by latent confounders\\
\noindent \rule{\columnwidth}{0.5mm}
 {\bf INPUT:} Data matrix $\X$ and a threshold $\alpha$
  \begin{enumerate}
  \item Given a $d$-dimensional random vector $\bfx$, a $d\times n$ data matrix of the random vector as $\X$ and a significance level $\alpha$, define $U$ as the set of variable indices of $\bfx$, {\it i.e.},   $\{1, \cdots, d\}$ and initialize an ordered list of variables $K_{head}:=\emptyset$ and $K_{tail}:=\emptyset$ and $m:=1$. $K_{head}$ and $K_{tail}$ denote the first $|K_{head}|$ variable indices and the last $|K_{tail}|$ variable indices respectively, where each of $|K_{head}|$ and $|K_{tail}|$ denotes the number of elements in the list. 
   \item Let $\Tilde{\bfx}:=\bfx$ and $\Tilde{\X}:=\X$ and find causal orders one by one from the top downward: \label{alg:uppers}
  \begin{enumerate}
   \item Do the following steps for all $j \in U \setminus K_{head}$:  \label{alg:findxm}
 Perform least squares regressions of $\tilde{x}_i$ on $\tilde{x}_j$ for all $i \in U \setminus K_{head}$ $(i \neq j)$ and compute the
residual vectors $\Tilde{\bfr}^{(j)}$ and the residual matrix $\Tilde{\R}^{(j)}$. 
Then, find a variable $\tilde{x}_m$ that is most independent of its residuals:
     \begin{equation} \label{alg:eq1}
      \Tilde{x}_m=\arg \max_{j \in U \setminus K_{head}}P_{Fisher}(\Tilde{x}_j,\Tilde{\bfr}^{(j)}), 
     \end{equation}
     where $P_{Fisher}(\Tilde{x}_j,\Tilde{\bfr}^{(j)})$ is the $p$-value of the test statistic defined as \\ 
    $-2\sum_i \log\{P_{H}(\Tilde{x}_j,\Tilde{r}_i^{(j)})\}$, where $P_{H}(\Tilde{x}_j,\Tilde{r}_i^{(j)})$ is the $p$-value of the HSIC.
   \item Go to Step~\ref{alg:unders} if $P_{Fisher}(\Tilde{x}_m, \Tilde{\bfr}^{(m)}) < \alpha$, {\it i.e.}, all independencies are rejected. \label{alg:rejec1}
   \item \label{step:appendhead} Append $m$ to the end of $K_{head}$ and let $\Tilde{\bfx}:=\Tilde{\bfr}^{(m)}$ and $\Tilde{\X}:=\Tilde{\R}^{(m)}$. 
   If $|K_{head}|=d-1$, append the remaining variable index to the end of $K_{head}$ and terminate. 
   Otherwise, go back to Step~(\ref{alg:findxm}). 
  \end{enumerate}
 \item If $|K_{head}| < d-2$, let $\bfx'=\bfx$ and $\X' =\X$ and $U':=U \setminus K_{head}$ and find causal orders one by one from the bottom upward \footnote{We do not examine remaining two variables in this step since it is already implied in Step~\ref{alg:uppers} that some latent confounders exist. If there were no latent confounders between the remaining two, their causal orders would have already been estimated in Step~\ref{alg:uppers}.}:
  \label{alg:unders}
  \begin{enumerate}
   \item Do the following steps for all $j \in U' \setminus K_{tail}$:  \label{alg:findxm2}
 Collect all the variables except $x'_j$ in a vector $\bfx'_{(-j)}$.  Perform least squares regressions of $x'_j$ on $\bfx'_{(-j)}$ and compute the residual ${r'}_j^{(-j)}$.  
    Then, find such a variable $x'_m$ that its regressors and its residual are most independent: 
     \begin{equation} \label{alg:eq2}
      x'_m=\arg \max_{j \in U' \setminus K_{tail}}P_{Fisher}(\bfx'_{(-j)},{r'}_j^{(-j)}).
     \end{equation}
  \item Terminate if $P_{Fisher}(\bfx'_{(-m)},{r'}_m^{(-m)}) < \alpha$, {\it i.e.}, all independencies are rejected. \label{alg:rejec2}
  \item \label{step:append} Append $m$ to the top of $K_{tail}$ and let $\bfx'=\bfx'_{(-m)}$，$\X'=\X'_{(-m)}$. 
  Terminate~\footnotemark[4] if $|U' \setminus K_{tail}|<3 $ and otherwise go back to Step~(\ref{alg:findxm2}). 
  \end{enumerate}
  \item Estimate a causal ordering matrix $\C$ based on $K_{head}$ and $K_{tail}$ as follows.   
 Estimate $c_{ij}$ by -1, {\it i.e.}, $k(i)<k(j)$ in either of the following cases: 
  i) $i$ is earlier than $j$ in $K_{head}$; 
  ii) $i$ is earlier than $j$ in $K_{tail}$;
  iii) $i$ is in $K_{head}$ and $j$ is in $K_{tail}$;
  iv) $i$ is in $K_{head}$ and $j$ is neither $K_{head}$ nor $K_{tail}$.   
  Estimate $c_{ij}$ by 1, {\it i.e.}, $k(i)>k(j)$ in either of the following cases: 
  i) $i$ is later than $j$ in $K_{head}$; 
  ii) $i$ is later than $j$ in $K_{tail}$;
  iii) $i$ is in $K_{tail}$ and $j$ is in $K_{head}$;
 iv) $i$ is in $K_{tail}$ and $j$ is neither $K_{head}$ nor $K_{tail}$.   
  Estimate $c_{ij}$ by 0, {\it i.e.}, the ordering is unknown if $i$ and $j$ are neither in $K_{tail}$ nor $K_{head}$. 
  Note that causal orders of variables that are not in $K_{head}$ or $K_{tail}$ are no later than any in $K_{tail}$ and no earlier than any in $K_{head}$.
  \end{enumerate}
 {\bf OUTPUT:} Ordered lists $K_{head}$ and $K_{tail}$ and a causal ordering matrix $\C$ 
\noindent \\
  \rule{\columnwidth}{0.5mm}

\begin{figure}[tb]
  \begin{center}
   \includegraphics[width=0.25\textwidth]{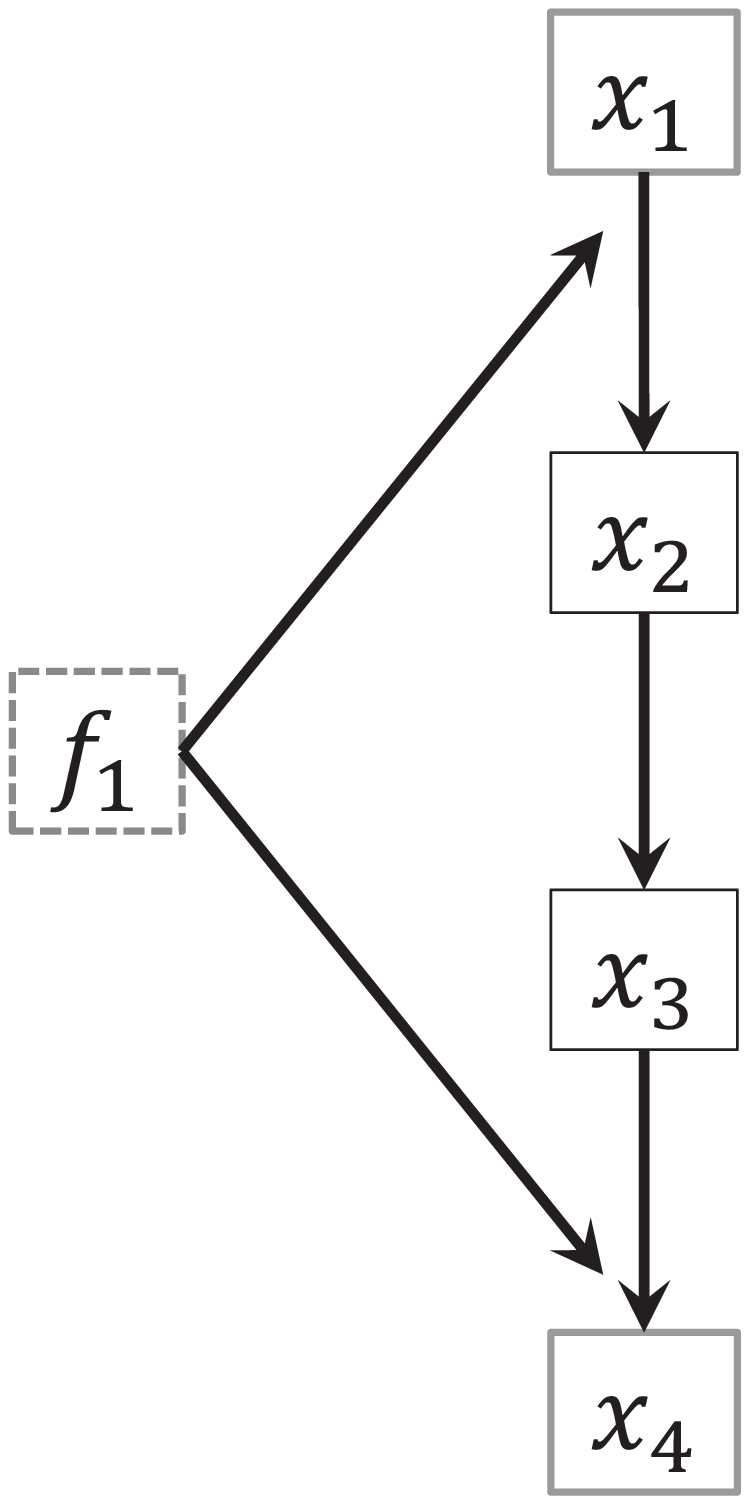}
   \hspace*{15mm}
   \includegraphics[width=0.25\textwidth]{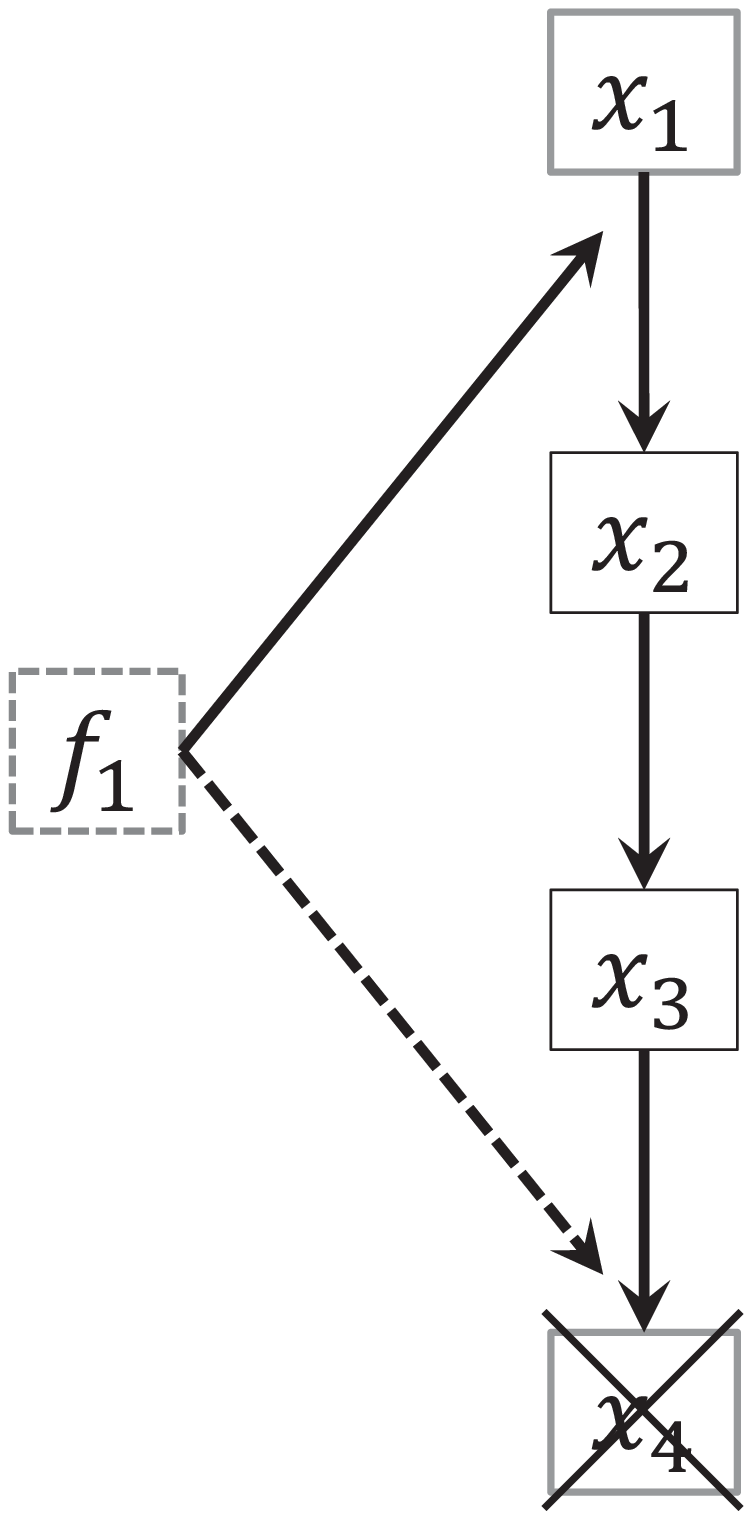}
   \caption{Left: An example graph where Algorithm~1 finds no causal orders. The $f_1$ is a latent confounder that affects $x_1$ and $x_4$. Right: Algorithm~1 finds the causal orders of $x_1$, $x_2$ and $x_3$ if $x_4$ is omitted and only $x_1$, $x_2$ and $x_3$ are analyzed. }
   \label{fig:dagexample}
  \end{center}
\end{figure}

\subsection{A new estimation algorithm robust against latent confounders}
Algorithm~1 outputs no causal orders in cases where exogenous variables and sink variables as in Lemmas~\ref{lemma1} and \ref{lemma2} do not exist. 
For example, in the left of Fig.~\ref{fig:dagexample}, there is no such exogenous variable or sink variable that is not affected by any latent confounder since the latent confounder $f_1$ affects the exogenous variable $x_1$ and the sink variable $x_4$. 
Therefore, Algorithm~1 would not find any causal orders. 
However, if we omit $x_4$ as in the right of Fig.~\ref{fig:dagexample} and apply Algorithm~1 on the remaining $x_1, x_2, x_3$ only, it will find all the causal orders of $x_1, x_2, x_3$ since $f_1$ does not affect any two of $x_1, x_2, x_3$ and is no longer a latent confounder. 
The same idea applies to the case that $x_1$ is omitted. 

Thus, we propose applying Algorithm~1 on every subset of variables with the size larger than one. 
This enables learning more causal orders than analyzing the whole set of variables if a subset of variables has exogenous variables or sink variables that are not affected by latent confounders. 
In practice, Algorithm~1 could give inconsistent causal orderings between a pair of variables for different subsets of variables because of estimation errors. 
To manage possible inconsistencies in the many causal orderings thus
estimated, we rank the obtained causal ordering matrices by
plausibility based on the statistical significances (this will be
defined below). Then, considering any pair of two variables, we use
the causal ordering given by the causal ordering matrix which has the
highest plausibility and does contain an estimated causal ordering
({\it i.e.}, the ordering was not considered unknown) between those two
variables.

We evaluate the plausibility of every causal ordering matrix by the $p$-value of the test statistic created based on Fisher's method combining all the $p$-values computed to estimate the causal orders $K_{head}$ and $K_{tail}$ in Algorithm~1. A higher $p$-value can be considered to be more plausible. 
The test statistic is computed based on $\X$, $K_{head}$ and $K_{tail}$ as follows:
  \begin{eqnarray}
 -2( \sum_{m \in K_{head}}\sum_{i: k(i)>k(m)} \log\{P_{H}(\Tilde{x}_m,\Tilde{r}_i^{(m)})\}
  + \sum_{m \in K_{tail}}\sum_{i: k(i)<k(m)} \log\{P_{H}(x'_{i},{r'}_m^{(-m)})\}), \label{eq:fit}
  \end{eqnarray}
where $P_{H}(\Tilde{x}_m,\Tilde{r}_i^{(m)})$ and $P_{H}(x'_{i},{r'}_m^{(-m)})$ are the $p$-values computed to estimate ordered lists $K_{head}$ and $K_{tail}$ in Algorithm~1. 

Thus, the estimation consists of the following steps:
\noindent \\
  \rule{\columnwidth}{0.5mm}
 Algorithm~2: Applying Algorithm~1 on every subset of variables and merging results\\
 \noindent \rule{\columnwidth}{0.5mm}
 {\bf INPUT:} Data matrix $\X$ and a threshold $\alpha$
 \begin{enumerate}
 \item Take all the $l$-combinations of variable indices $\{1,\cdots, d\}$ for $l=2,\cdots,d$. Denote the subsets of variable indices by $U_{subset}^{(s)}$ $(s =1,\cdots,S)$ and the corresponding data matrices by $\X_{subset}^{(s)}$ $(s=1,\cdots, S)$, where $S$ is the number of the subsets.  
 \item Apply Algorithm~1 on $\X_{subset}^{(s)}$ using the threshold $\alpha$ to estimate $K_{head}^{(s)}$, $K_{tail}^{(s)}$ and $\C^{(s)}$ for all $s \in \{1,\cdots,S\}$, where $K_{head}^{(s)}$ and $K_{tail}^{(s)}$ are ordered lists of Subset $U_{subset}^{(s)}$ and $\C^{(s)}$ is a causal ordering matrix of Subset $U_{subset}^{(s)}$.
  \item Compute the $p$-value of the test statistic in Eq.~(\ref{eq:fit}) to evaluate the plausibility of $\C^{(s)}$ for all $s \in \{1,\cdots,S\}$.
 \item \label{step:createc} Estimate every element $c_{ij}$ ($i\neq j$) of a causal ordering matrix $\C$ by the causal ordering between $x_i$  and $x_j$ of the causal ordering matrix that has the highest plausibility and does contain an estimated causal ordering between $x_i$ and $x_j$, that is, $k(i)<k(j)$ or $k(j)<k(i)$.
  \end{enumerate}
 {\bf OUTPUT:} A causal ordering matrix $\C$
 \noindent \\
  \rule{\columnwidth}{0.5mm}

Algorithm~2 is a brute force approach since it applies Algorithm~1 on {\it every} subset (parcel) of variables. 
We could alleviate the computational load by first applying Algorithm~1 on the whole set of variables and then applying Algorithm~2 on the remaining variables whose causal orders have not been estimated after the effects of estimated exogenous variables are removed by regression. 
Thus, we finally propose the following algorithm called ParceLiNGAM: 
\noindent \\
  \rule{\columnwidth}{0.5mm}
 Algorithm~3:  The ParceLiNGAM algorithm
 \noindent \\
  \rule{\columnwidth}{0.5mm}
\\ {\bf INPUT:} Data matrix $\X$ and a threshold $\alpha$
 \begin{enumerate}
     \item Given a $d$-dimensional random vector $\bfx$ and a $d\times n$ data matrix of the random vector as $\X$, define $U$ as the set of variable indices of $\bfx$, {\it i.e.}, $\{1, \cdots, d\}$.
      initialize a $d \times d$ causal ordering matrix $\C$ by the zero matrix. 
     \item Apply Algorithm~1 on $\X$ using the threshold $\alpha$ to estimate $K_{head}$ and $K_{tail}$ and update $\C$. 
      \item Let $U_{res} := U\setminus (K_{head }\bigcup K_{tail})$. 
      Denote by $\C_{res}$ the corresponding causal ordering matrix. 
      Denote by $|U_{res}|$ the number of elements in $U_{res}$. Go to Step~\ref{step:estb} if $|U_{res}|\leq 2$.  
 \item     Collect variables $x_j$ with $j \in U_{res}$ in a vector $\bfx_{res}$.  
Collect variables $x_j$ with $j \in K_{head}$ in a vector $\bfx_{head}$.  
 Perform least squares regressions of $\bfx_{head}$ on the $i$-th element of $\bfx_{res}$ for all $i \in U_{res}$ and collect the residuals in the residual matrix $\R_{res}$ whose $i$-th row is given by the residuals regressed on $x_i$.
     \item Apply Algorithm~2 on $\R_{res}$ using the threshold $\alpha$ to estimate $\C_{res}$. 
    Replace every $c_{ij}$ ($i\neq j$) of $\C$ by the corresponding element of $\C_{res}$ if $c_{ij}$ is zero and the corresponding element of $\C_{res}$ is 1 or -1.
 \item  \label{step:estb} Estimate connection strengths $b_{ij}$ if all the non-descendants of $x_i$ are estimated,  {\it i.e.}, the $i$-th row of $\C$ has no zero. 
 This can be done by doing multiple regression of  $x_i$ on all of its non-descendants $x_j$ with $k(j)<k(i)$. 
 \end{enumerate}
 {\bf OUTPUT:} A causal ordering matrix $\C$ and a set of estimated connection strength $b_{ij}$. 
\noindent \\
  \rule{\columnwidth}{0.5mm}
  In cases of no latent confounders, Algorithm~3 is essentially equivalent to \\
DirectLiNGAM~\cite{Shimizu11JMLR}. 
Matlab codes for performing Algorithm~3 are available at \url{http://www.ar.sanken.osaka-u.ac.jp/~sshimizu/code/Plingamcode.html}.

%
%
%

\begin{figure*}
  \begin{center}
   \hspace*{-11mm}
 \begin{minipage}{0.54\textwidth}
   \centering
   \includegraphics[width=\textwidth]{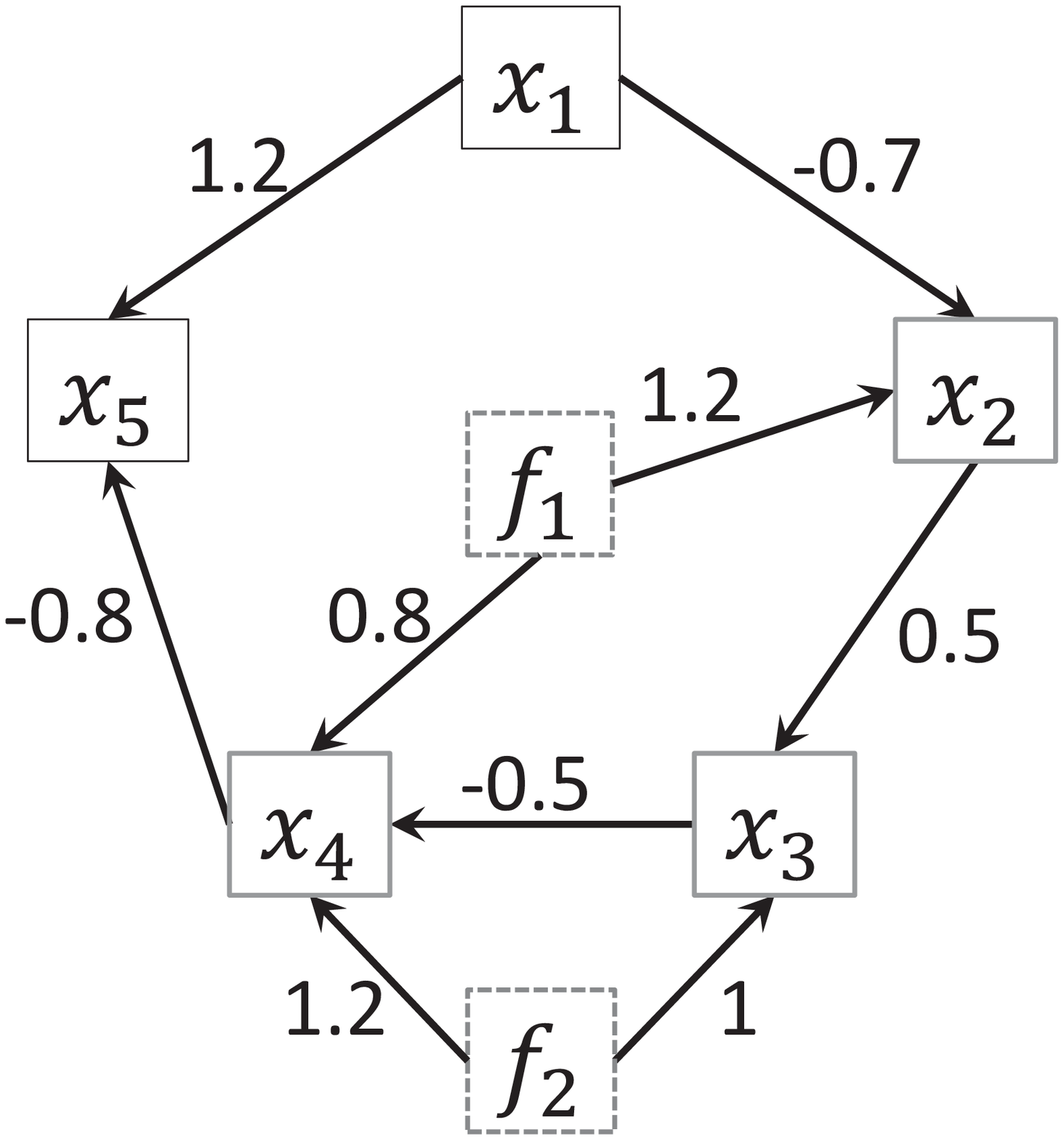}
   \caption{5 variable network}
   \label{fig:sim5dim}
 \end{minipage}
 \hspace*{-6mm}
 \begin{minipage}{0.54\textwidth}
   \centering
   \includegraphics[width=\textwidth]{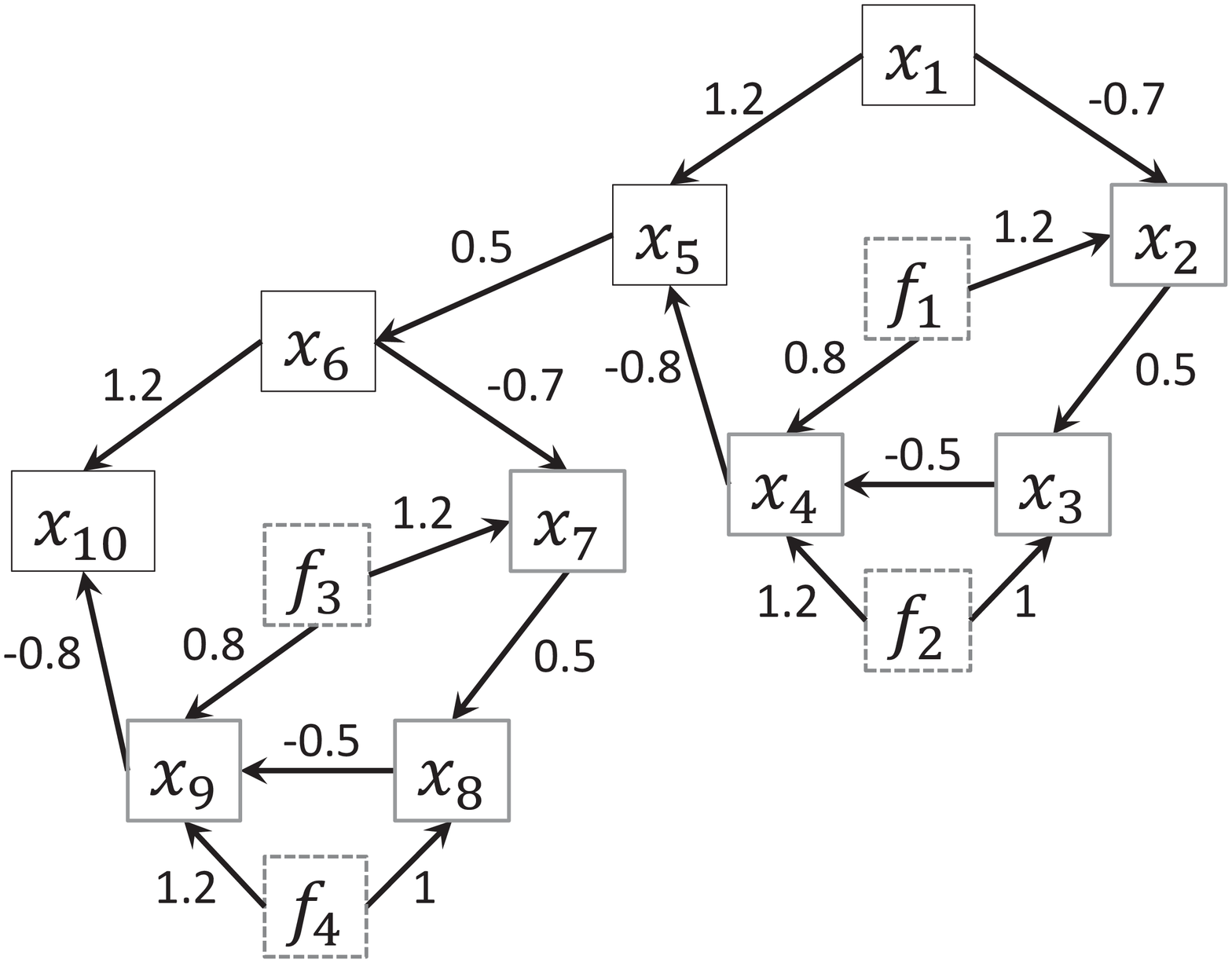}
   \caption{10 variable network}
   \label{fig:sim10dim}
 \end{minipage}
 \begin{minipage}{0.84\textwidth}
   \centering
   \includegraphics[width=\textwidth]{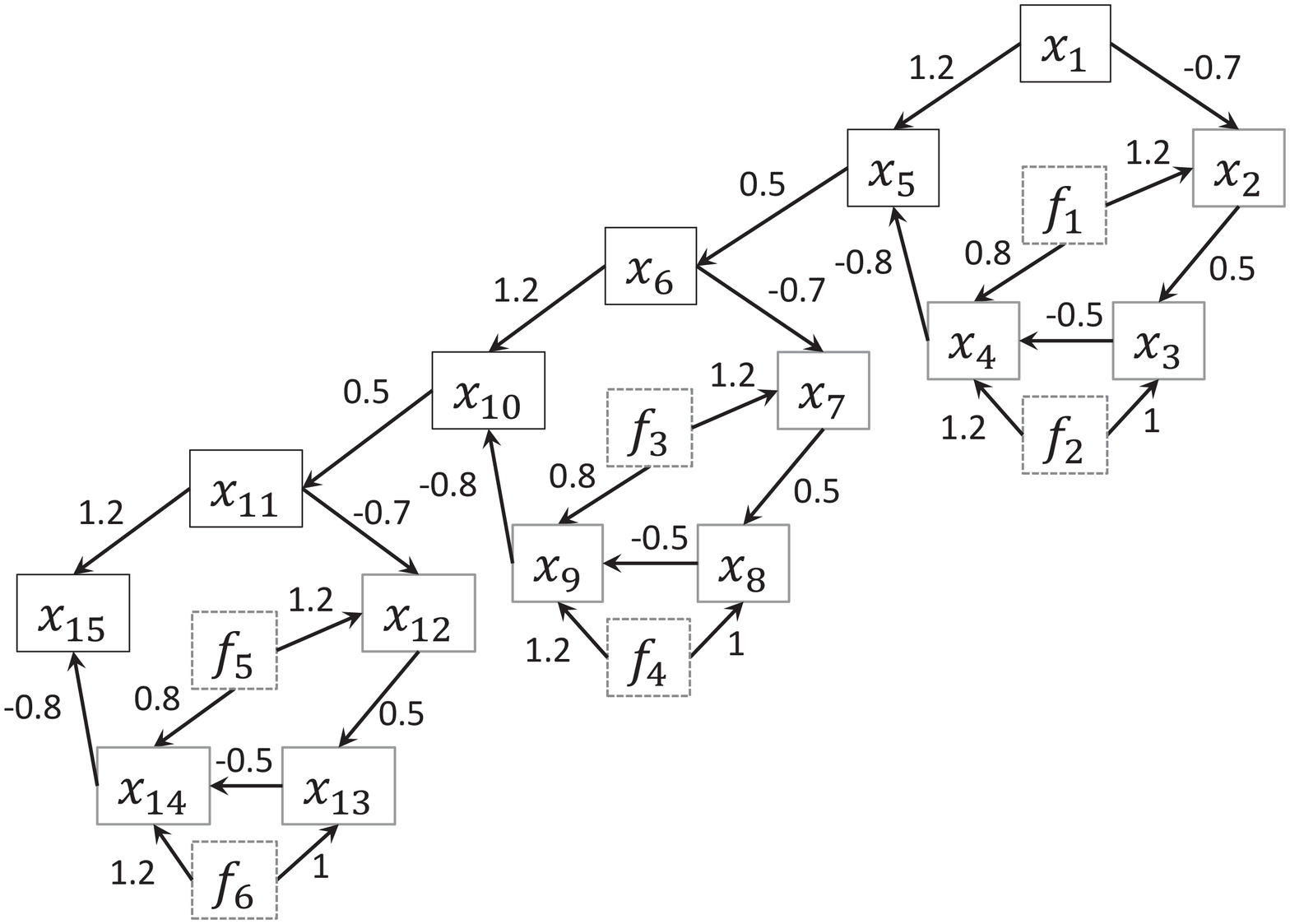}
   \caption{15 variable network}
   \label{fig:sim15dim}
 \end{minipage}
  \end{center}
\end{figure*}

\section{Experiments on artificial data}\label{sec:exp}
We compared our method with two estimation methods for LiNGAM in Eq.~(\ref{eq:model2}) called ICA-LiNGAM \cite{Shimizu06JMLR} and DirectLiNGAM \cite{Shimizu11JMLR} that do not allow latent confounders and an estimation method for latent variable LiNGAM in Eq.~(\ref{eq:lvlingam}) called Pairwise~LvLiNGAM \cite{Entner10AMBN}. 
If there are no latent confounders, all the methods should estimate correct causal orders for large enough sample sizes. 
The numbers of variables were 5, 10, and 15, and the sample sizes tested were 500, 1000, and 1500. 
The original networks used were shown in Fig.~\ref{fig:sim5dim} to Fig.~\ref{fig:sim15dim}.
The $e_1$, $e_4$, $e_7$, $e_{10}$, $e_{13}$, $f_1$ and $f_4$ followed a multimodal asymmetric mixture of two Gaussians,
$e_2$, $e_5$, $e_8$, $e_{11}$, $e_{14}$, $f_2$ and $f_5$ followed a double exponential distribution, 
and $e_3$, $e_6$, $e_9$, $e_{12}$, $e_{15}$, $f_3$ and $f_6$ followed a multimodal symmetric mixture of two Gaussians. 
The variances of the $e_i$ were set so that ${\rm var}(e_i)/{\rm var}(x_i)$$=$$1/2$. 
We permuted the variables according to a random ordering.
The number of trials was 100. The significance level $\alpha$ was 0.05. 

First, to evaluate performance of estimating causal orders $k(i)$, we computed the percentage of correctly estimated causal orders among estimated causal orders between two variables (Precision) and the percentage of correctly estimated causal orders among actual causal orders between two variables (Recall). 
We also computed the F-measure defined as $2 \times {\rm Precision} \times {\rm Recall} / ({\rm Precision} + {\rm Recall})$, which is the harmonic mean of Precision and Recall. 
The reason why only pairwise causal orders were evaluated was that Pairwise LvLiNGAM only estimates causal orders of two variables unlike our method and DirectLiNGAM.
Tables~\ref{sim:precision}, \ref{sim:recall} and \ref{sim:fmeasure} show the results. 
Regarding recalls and F-measures, the maximal performances when no statistical errors occur are also shown in the right-most columns. 
For example in Fig.~\ref{fig:sim5dim}, Pairwise~LvLiNGAM can find all the causal orderings except $k(2)< k(4)$, $k(2)<k(5)$, $k(3)<k(4)$ and $k(3)<k(5)$. 
ParceLiNGAM further can find $k(2)<k(5)$ and $k(3)<k(5)$ since it estimates causal orderings between more than two variables. 
In some cases, the empirical recalls and F-measures were higher than their maximal performances. This is because causal orders of some variables that are affected by latent confounders happened to be correctly estimated. 
Regarding precisions and F-measures, our method ParceLiNGAM worked best for all the conditions. 
Regarding recalls, ParceLiNGAM worked best for most conditions and was the second-best but comparable to the best method DirectLiNGAM for the other conditions. 

Next, to evaluate the performance in estimating connection strengths $b_{ij}$, 
we computed the root mean square errors between true connection strengths and estimated ones. 
Note that Pairwise~LvLiNGAM does not estimate $b_{ij}$. 
Table~\ref{sim:rmse} show the results. 
Our method was most accurate for all the conditions. 

Table~\ref{sim:time} shows average computation times. 
The amount out computation of our 
ParceLiNGAM was larger than the other methods when the sample size was increased. 
However, its amount of computation can be considered to be still tractable. 
For larger numbers of variables, we would need to select a subset of variables to decrease the number of variables to be analyzed. However, this selection does not bias results of our method since it allows latent confounders. 



\begin{table}[htbp]
\begin{center}
\caption{Precisions}
 \scriptsize
 \begin{tabular}{ll|rrr}
     &   & \multicolumn{3}{c}{Sample size} \\
     &   & 500 & 1000 & 1500 \\ \hline \hline
 ParceLiNGAM  & dim.=5 & 1.0 & 1.0 & 1.0 \\
     & dim.=10 & 0.81	& 0.88	& 0.93	\\
     & dim.=15 & 0.81	& 0.89	& 0.92	\\ \hline
 PairwiseLvLiNGAM& dim.=5 & 0.87 & 0.94 & 0.94 \\
     & dim.=10 & 0.75 & 0.79 & 0.81 \\
     & dim.=15 & 0.67 & 0.76 & 0.75 \\ \hline
 DirectLiNGAM & dim.=5 & 0.82 & 0.88 & 0.85 \\
     & dim.=10 & 0.59 & 0.71 & 0.73 \\
     & dim.=15 & 0.78 & 0.80 & 0.82 \\ \hline
 ICA-LiNGAM  & dim.=5 & 0.80 & 0.75 & 0.76 \\
     & dim.=10 & 0.62 & 0.62 & 0.58 \\
     & dim.=15 & 0.58 & 0.59 & 0.58 
 \end{tabular}
\label{sim:precision}
\end{center}
\end{table}%

\begin{table}[htbp]
\begin{center}
\caption{Recalls}
 \scriptsize
 \begin{tabular}{ll|rrr|c}
     &   & \multicolumn{3}{c|}{Sample size} & Max. performance \\
     &   & 500 & 1000 & 1500 &  \\ \hline \hline
 ParceLiNGAM  & dim.=5 & 0.86	& 0.82	& 0.80 & 0.80(8/10)  \\
     & dim.=10 & 0.79	& 0.85	& 0.91 & 0.91(41/45)  \\
     & dim.=15 & 0.80	& 0.87	& 0.89 & 0.94(99/105)  \\ \hline
 PairwiseLvLiNGAM& dim.=5 & 0.65 & 0.62 & 0.59 & 0.60(6/10)  \\
     & dim.=10 & 0.50 & 0.55 & 0.54 & 0.49(22/45)  \\
     & dim.=15 & 0.39 & 0.45 & 0.43 & 0.46(48/105)  \\ \hline
 DirectLiNGAM & dim.=5 & 0.82 & 0.88 & 0.85 & - \\
     & dim.=10 & 0.59 & 0.71 & 0.73 & - \\
     & dim.=15 & 0.78 & 0.80 & 0.82 & - \\ \hline
 ICA-LiNGAM  & dim.=5 & 0.80 & 0.75 & 0.76 & - \\
     & dim.=10 & 0.62 & 0.62 & 0.58 & - \\
     & dim.=15 & 0.58 & 0.59 & 0.58 & - 
 \end{tabular}
\label{sim:recall}
\end{center}
\end{table}%

\begin{table}[htbp]
\begin{center}
\caption{F-measures}
 \scriptsize
 \begin{tabular}{ll|rrr|c}
     &   & \multicolumn{3}{c|}{Sample size} & Max. performance \\
     &   & 500 & 1000 & 1500 &  \\ \hline \hline
 ParceLiNGAM  & dim.=5 & 0.92	& 0.90	& 0.89 & 0.89 \\
     & dim.=10 & 0.80	& 0.86	& 0.92 & 0.95 \\
     & dim.=15 & 0.81	& 0.88	& 0.90 & 0.97 \\ \hline
 PairwiseLvLiNGAM& dim.=5 & 0.75 & 0.75 & 0.72 & 0.75 \\
     & dim.=10 & 0.60 & 0.65 & 0.65 & 0.66 \\
     & dim.=15 & 0.49 & 0.56 & 0.54 & 0.63  \\ \hline
 DirectLiNGAM & dim.=5 & 0.82 & 0.88 & 0.85 & - \\
     & dim.=10 & 0.59 & 0.71 & 0.73 & - \\
     & dim.=15 & 0.78 & 0.80 & 0.82 & - \\ \hline
 ICA-LiNGAM  & dim.=5 & 0.80 & 0.75 & 0.76 & - \\
     & dim.=10 & 0.62 & 0.62 & 0.58 & - \\
     & dim.=15 & 0.58 & 0.59 & 0.58 & - 
 \end{tabular}
\label{sim:fmeasure}
\end{center}
\end{table}%

\begin{table}[htbp]
\begin{center}
\caption{Root Mean Square Errors}
 \scriptsize
 \begin{tabular}{ll|rrrr}
     &   & \multicolumn{3}{c}{Sample size} \\
     &   & 500 & 1000 & 1500 \\ \hline \hline
 ParceLiNGAM  & dim.=5	& 0.030	& 0.020	& 0.016	\\
     & dim.=10 & 0.078	& 0.060	& 0.052	\\
     & dim.=15 & 0.083	& 0.046	& 0.031	\\ \hline
 DirectLiNGAM & dim.=5 & 0.22 & 0.16 & 0.18 \\
     & dim.=10 & 0.16 & 0.083 & 0.089 \\
     & dim.=15 & 0.096 & 0.074 & 0.070 \\ \hline
 ICA-LiNGAM  & dim.=5 & 0.11 & 0.11 & 0.10 \\
     & dim.=10 & 0.16 & 0.15 & 0.15 \\
     & dim.=15 & 0.16 & 0.14 & 0.13 
 \end{tabular}
\label{sim:rmse}
\end{center}
\end{table}%

\begin{table}[htbp]
\begin{center}
\caption{Computational Times}
 \scriptsize
 \begin{tabular}{ll|rrr}
     &   & \multicolumn{3}{c}{Sample size} \\
     &   & 500 & 1000 & 1500 \\ \hline \hline
 ParceLiNGAM  & dim.=5	& 0.66 sec.	& 1.7 sec.	& 4.4 sec.	\\
     & dim.=10 & 10 sec.	& 1.5 min.	& 8.1 min.	\\
     & dim.=15 & 8.5 min.	& 5.3 hrs.	& 19 hrs.	\\ \hline
 PairwiseLvLiNGAM& dim.=5 & 0.64 sec. & 2.6 sec. & 7.0 sec. \\
     & dim.=10 & 2.8 sec. & 12 sec. & 30 sec. \\
     & dim.=15 & 6.6 sec. & 29 sec. & 74 sec. \\ \hline
 DirectLiNGAM & dim.=5 & 0.23 sec. & 0.84 sec. & 1.2 sec. \\
     & dim.=10 & 1.7 sec. & 7.3 sec. & 11 sec. \\
     & dim.=15 & 6.4 sec. & 29 sec. & 44 sec. \\ \hline
 ICA-LiNGAM  & dim.=5 & 0.12 sec. & 0.051 sec.& 0.047 sec. \\
     & dim.=10 & 0.34 sec. & 0.18 sec. & 0.10 sec. \\
     & dim.=15 & 0.81 sec. & 0.68 sec. & 0.53 sec. 
 \end{tabular}
\label{sim:time}
\end{center}
\end{table}%

\section{Experiments on simulated fMRI data}\label{sec:real}
\begin{figure*}
 \begin{center}
  \includegraphics[width=0.75\textwidth]{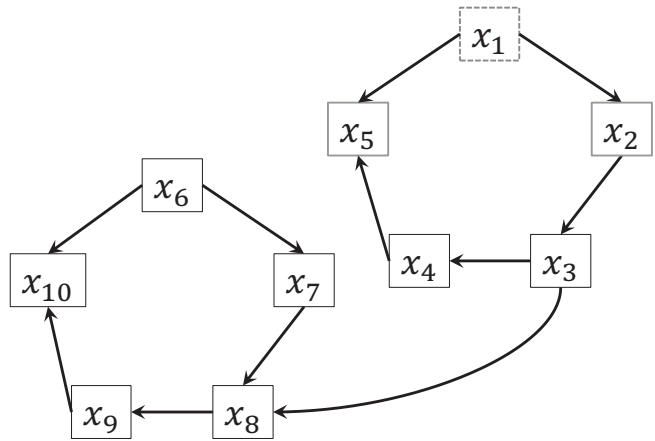}
  \caption{The network used in the simulated fMRI experiments. We omitted $x_1$ to create a latent confounder. 
  }
  \label{smith:network}
 \end{center}
\end{figure*}

Finally, we tested our method on simulated functional magnetic resonance imaging (fMRI) data generated in \cite{Smith11NI} based on a well-known mathematical brain model called the dynamic causal modeling \cite{Friston03NI}. 
We used Simulation~2 data and Simulation~6 data. 
Both datasets consisted of 10 variables whose causal structure is shown in Fig.~\ref{smith:network}. 
The session durations were 10 minutes (200 time points) and 60 minutes (1200 time points), respectively. 
We also created a dataset of 30 minutes (600 time points) by taking the first half of Simulation~6 data. 
Although these data are time-series, we did not add lag-based approaches including vector autoregressive models into comparison as in \cite{Hyva13JMLR} since it was shown by \cite{Smith11NI} that lag-based methods worked poorly on these Simulation 2 data and Simulation 6 data. 

For each of the three different duration settings, we gave the 50 datasets (one by one) to
ParceLiNGAM, PairwiseLvLiNGAM, DirectLiNGAM and ICA-LiNGAM after omitting $x_1$ to create a latent confounder and randomly permuting the other variables.
Table~\ref{smith:order} shows the precision, recalls, and F-measures of causal orders. 
Regarding precisions, we excluded such variable pairs $x_i$ and $x_j$ that one has no directed path to the other, e.g., $x_2$ and $x_6$, since both $k(i)<k(j)$ and $k(i)>k(j)$ are correct. 
This was because estimation of causal directions is the main topic of this paper.   
The significance level $\alpha$ was 0.05. 
For all of the cases, ParceLiNGAM worked better than the others. 

\begin{table}[htbp]
\begin{center}
\caption{Results on simulated fMRI data}
 \scriptsize
 \begin{tabular}{ll|ccc}
      &   & sim2 (10 min.) & sim6 (30 min.) & sim6 (60 min.) \\ \hline \hline
 ParceLiNGAM  & Precision & 0.54	& 0.56	& 0.60	\\
      & Recall & 0.53	& 0.55	& 0.58	\\
      & F-measure & 0.53	& 0.55	& 0.59	\\ \hline
 PairwiseLvLiNGAM & Precision & 0.31 & 0.25 & 0.24 \\
      & Recall & 0.22 & 0.15 & 0.14 \\
      & F-measure & 0.26 & 0.19 & 0.18 \\ \hline
 DirectLiNGAM  & Precision & 0.50 & 0.51 & 0.45 \\
      & Recall & 0.50 & 0.51 & 0.45 \\
      & F-measure & 0.50 & 0.51 & 0.45 \\ \hline
 ICA-LiNGAM   & Precision & 0.49 & 0.47 & 0.47 \\
      & Recall & 0.49 & 0.47 & 0.47 \\
      & F-measure & 0.49 & 0.47 & 0.47
 \end{tabular}
\label{smith:order}
\end{center}
\end{table}

\section{Conclusions}\label{sec:conc}
We proposed a new algorithm for learning causal orders, which is robust against latent confounders. 
In experiments on artificial data and simulated fMRI data, our methods learned more causal orders correctly than existing methods. 
An important problem for future research is to develop computationally more efficient algorithms. One approach might be to develop a divide-and-conquer algorithm that divides variables into subsets with moderate numbers of variables and integrates the estimation results on the subsets. 

\subsubsection*{Acknowledgments.}
S.S and T.W. were supported by KAKENHI \#24700275 and \#22300054. 
We thank Patrik Hoyer and Doris Entner for helpful comments.

\bibliography{shimizu12a,tsuika_bun}{}
\bibliographystyle{apacite}

\section*{Appendix}\label{appendix}
We first give Darmois-Skitovitch theorem \cite{Darmois1953,Skitovitch53}: 
\begin{thm}[Darmois-Skitovitch theorem (D-S theorem)]\label{thm1}
Define two random variables $y_1$ and $y_2$ as linear combinations of independent random variables $s_i$($i$$=$1, $\cdots$, $q$): $y_1 = \sum_{i=1}^q \alpha_is_i$, $y_2 = \sum_{i=1}^q \beta_i s_i$.
Then, if $y_1$ and $y_2$ are independent, all variables $s_j$ for which $\alpha_j\beta_j\neq0$ are Gaussian. \mbox{\hfill \qed}
\end{thm}
In other words, this theorem means that if there exists a non-Gaussian $s_j$ for which $\alpha_j\beta_j$$\neq$$0$, $y_1$ and $y_2$ are dependent. 


\subsubsection*{Proof of Lemma~\ref{lemma1}}
i) Assume that $x_j$ has at least one parent observed variable or latent confounder. 
Let $P_j$ denote the set of the parent variables of $x_j$. 
Then one can write $x_j$$=$$ \sum_{p_h \in P_j} w_{jh} p_h$$+$$e_j$, where the parent variables $p_h$ are independent of $e_j$ and the coefficients $w_{jh}$ are non-zero. 
Let a vector $\x_{P_j}$ and a column vector $\bfw_{P_j}$ collect all the variables in $P_j$ and the corresponding connection strengths, respectively. 
Then, the covariances between $\x_{P_j}$ and $x_j$ are 
$E(\x_{P_j}x_j) = E\{\x_{P_j} (\bfw_{P_j}^T \x_{P_j}+e_j) \} 
 = E(\x_{P_j} \x_{P_j}^T)\bfw_{P_j}.  $ 
The covariance matrix $E(\x_{P_j} \x_{P_j}^T)$ is positive definite since the external influences and latent confounders are mutually independent and have  positive variances.
Thus, the covariance vector $E(\x_{P_j}x_j)=E(\x_{P_j} \x_{P_j}^T)\bfw_{P_j}$ above cannot equal the zero vector, 
and there must be at least one variable in $P_j$ with which $x_j$ covaries. 

i-a) Suppose that $x_i$ is a parent of $x_j$ in $P_j$ that covaries with $x_j$. 
For such $x_i$, we have 
\begin{eqnarray}
r_i^{(j)} &=& x_i - \frac{{\rm cov}( x_i, x_j )}{{\rm var}(x_j)}x_j  \\
 & =& x_i - \frac{{\rm cov}( x_i, x_j )}{{\rm var}(x_j)}(\sum_{p_h \in P_j} w_{jh} p_h + e_j) \\
 & =& \left\{1-\frac{w_{ji}{\rm cov}(x_i,x_j)}{{\rm var}(x_j)}\right\} x_i -\frac{{\rm cov}(x_i,x_j)}{{\rm var}(x_j)}\sum_{p_h\in P_j,  p_h\neq x_i}w_{jh}p_h \nonumber\\
& &    - \frac{{\rm cov}(x_i,x_j)}{{\rm var}(x_j)} e_j. 
\end{eqnarray}
Each of those parent variables (including $x_i$) in $P_j$ is a linear combination of external influences {\it other than} $e_j$ and latent confounders that are non-Gaussian and independent. 
Thus, the $r_i^{(j)}$ and $x_j$ can be written as linear combinations of non-Gaussian and independent external influences including $e_j$ and latent confounders.
Further, the coefficient of $e_j$ on $r_i^{(j)}$  is non-zero since ${\rm cov}(x_i,x_j)\neq 0$ aforementioned and that on $x_j$ is one by definition.  
These imply that $r_i^{(j)}$ and $x_j$ are dependent since $r_{i}^{(j)}$, $x_j$ and $e_j$ correspond to $y_1$, $y_2$, $s_j$ in D-S theorem, respectively.  

i-b) Next, suppose that $x_j$ has a latent confounder $f_k$ in $P_j$ that covaries with $x_j$. 
The latent confounder $f_k$ should have a non-zero coefficient on at least one other observed variable $x_i$. 
 Without loss of generality, it is enough to consider two observed variable cases that we only observe $x_i$ and $x_j$:
\begin{eqnarray}
x_i &=& b_{ij} x_j + \lambda_{ik} f_k + e_i + \sum_{h\neq k} \lambda_{ih} f_h \\
x_j &=& b_{ji} x_i + \lambda_{jk} f_k + e_j + \sum_{l\neq k} \lambda_{il} f_l, 
\end{eqnarray}
where $\lambda_{ik}$ and $\lambda_{jk}$ are non-zero since $f_k$ is a latent confounder of $x_i$ and $x_j$. 
Since the model is acyclic, $b_{ij}b_{ji}=0$. 

First, suppose that $b_{ij}$ is zero. 
Then, we have
\begin{eqnarray}
r_i^{(j)} &=& x_i - \frac{{\rm cov}( x_i, x_j )}{{\rm var}(x_j)}x_j  \\
&=& \{ \lambda_{ik}-\frac{{\rm cov}( x_i, x_j )}{{\rm var}(x_j)}(b_{ji}\lambda_{ik}+\lambda_{jk})\}f_k \nonumber \\
 & & + \{1-\frac{{\rm cov}( x_i, x_j )}{{\rm var}(x_j)}b_{ji}\}e_i - \frac{{\rm cov}( x_i, x_j )}{{\rm var}(x_j)}e_j + D_1, 
\end{eqnarray}
where $D_1$ is a linear combinations of non-Gaussian and independent latent confounders other than $f_k$. 
If ${\rm cov}( x_i, x_j )$ is zero, the coefficient of $f_k$ on $r_i^{(j)}$ is $\lambda_{ik}$ and is non-zero. 
If ${\rm cov}( x_i, x_j )$ is non-zero, the coefficient of $e_j$ on $r_i^{(j)}$ is $-\frac{{\rm cov}( x_i, x_j )}{{\rm var}(x_j)}$ and is non-zero. 
Thus, in both of the cases, $r_{i}^{(j)}$ and $x_j$ are dependent due to D-S theorem. 
Remember that the coefficient of $e_j$ on $x_j$ is one by definition. 

Next, suppose that $b_{ji}$ is zero. 
Then, we have
\begin{eqnarray}
r_i^{(j)} &=& x_i - \frac{{\rm cov}( x_i, x_j )}{{\rm var}(x_j)}x_j  \\
&=& \{(b_{ij}\lambda_{jk}+\lambda_{ik})-\frac{{\rm cov}( x_i, x_j )}{{\rm var}(x_j)}\lambda_{jk}\}f_k \nonumber \\
 & & + e_i + (b_{ij}- \frac{{\rm cov}( x_i, x_j )}{{\rm var}(x_j)})e_j + D_2, 
\end{eqnarray}
where $D_2$ is a linear combinations of non-Gaussian and independent latent confounders other than $f_k$. 
If ${\rm cov}( x_i, x_j )$ is zero and $b_{ij}$ is zero, the coefficient of $f_k$ on $r_i^{(j)}$ is $\lambda_{ik}$ and is non-zero. 
If ${\rm cov}( x_i, x_j )$ is zero and $b_{ij}$ is non-zero, the coefficient of $e_j$ on $r_i^{(j)}$ is $b_{ij}$ and is non-zero. 
If ${\rm cov}( x_i, x_j )$ is non-zero and $b_{ij}$ is zero, the coefficient of $e_j$ on $r_i^{(j)}$ is $-\frac{{\rm cov}( x_i, x_j )}{{\rm var}(x_j)}$ and is non-zero. 
If ${\rm cov}( x_i, x_j )$ is non-zero and $b_{ij}$ is non-zero, either of the followings holds: a) the coefficient of $e_j$ on $r_i^{(j)}$ is non-zero, that is, $b_{ij}\neq {\rm cov}( x_i, x_j )/{\rm var}(x_j)$ or b) the coefficient of $e_j$ on $r_i^{(j)}$ is zero and hence the coefficient of $f_k$ on $r_i^{(j)}$ is $\lambda_{ik}$ and is non-zero.  
Thus, in all of the cases, $r_{i}^{(j)}$ and $x_j$ are dependent due to D-S theorem.

 ii) The converse of contrapositive of i) is straightforward using the model definition.  
From i) and ii), the lemma is proven. 
\mbox{\hfill \bsquare}

\vspace{-3mm}

\subsubsection*{ Proof of Lemma~\ref{lemma2}}
i) Assume that a variable $x_j$  has at least one child observed variable or latent confounder. 
First, without loss of generality, one can write
\begin{eqnarray}
\bfx = 
\left[
\begin{array}{c}
x_j\\
\bfx_{(-j)}
\end{array}
\right]
&=& (\I-\B)^{-1} (\bLambda \bff + \bfe)
= \A (\bLambda \bff + \bfe)\\
&=&
\left[
\begin{array}{cc}
1 & \bfa_{j(-j)}^T\\
\bfa_{(-j)j} & \A_{(-j)}
\end{array}
\right]
\left[
\begin{array}{c}
{\pmb \lambda}_{j}^T\bff+e_j\\
\bLambda_{(-j)}\bff+\bfe_{(-j)}
\end{array}
\right], 
\end{eqnarray}
where each of $\A$ ($=(\I-\B)^{-1}$) and $\A_{(-j)}$ is invertible and can be permuted to be a lower triangular matrix with the diagonal elements being ones if the rows and columns are simultaneously permuted according to the causal ordering $k(i)$. The same applies to the inverse of $\A$: 
\begin{eqnarray}
\A^{-1} &=& 
\left[
\begin{array}{cc}
(1-\bfa_{j(-j)}^T\A_{(-j)}^{-1}\bfa_{(-j)j})^{-1} & -\bfa_{j(-j)}^T\D^{-1} \\
-\D^{-1}\bfa_{(-j)j} & \D^{-1} 
\end{array}
\right],
\end{eqnarray}
where $\D=\A_{(-j)}-\bfa_{(-j)j}\bfa_{j(-j)}^T$.
Thus, $1-\bfa_{j(-j)}^T\A_{(-j)}^{-1}\bfa_{(-j)j}=1$. 

Then, 
\begin{eqnarray}
r_{j}^{(-j)} &=& x_j - \bsigma^T_{(-j)j} \Sigma_{(-j)}^{-1} \bfx_{(-j)}\\
 &=& {\pmb \lambda}_j^T\bff + e_j + \bfa^T_{j(-j)}(\bLambda_{(-j)}\bff + \bfe_{(-j)})\nonumber \\
 & &- \bsigma^T_{(-j)j} \Sigma_{(-j)}^{-1} \{\bfa_{(-j)j}({\pmb \lambda}_{j}^T\bff+e_j)+\A_{(-j)}(\bLambda_{(-j)}\bff+\bfe_{(-j)})\\
&=& \{{\pmb \lambda}_j^T+\bfa_{j(-j)}^T\bLambda_{(-j)}- \bsigma^T_{(-j)j} \Sigma_{(-j)}^{-1}(\bfa_{(-j)j}{\pmb \lambda}_j^T + \A_{(-j)}\bLambda_{(-j)})\}\bff \nonumber\\
& & \hspace{-2.5mm} +\{1- \bsigma^T_{(-j)j} \Sigma_{(-j)}^{-1} \bfa_{(-j)j}\}e_j  + \{ \bfa_{j(-j)}^T - \bsigma^T_{(-j)j} \Sigma_{(-j)}^{-1} \A_{(-j)} \}\bfe_{(-j)}. \label{eq:prf_r}
\end{eqnarray}
In Eq.(\ref{eq:prf_r}), if $\bfa_{j(-j)}^T - \bsigma^T_{(-j)j} \Sigma_{(-j)}^{-1} \A_{(-j)} = {\bf 0}^T$, then we have
\begin{eqnarray}
r_{j}^{(-j)} &=& \{{\pmb \lambda}_j^T(1-\bfa_{j(-j)}^T \A_{(-j)}^{-1}\bfa_{(-j)j})\}\bff
 +\{1- \bfa_{j(-j)}^T \A_{(-j)}^{-1} \bfa_{(-j)j}\}e_j \\
 &=& {\pmb \lambda}_j^T\bff +e_j. \label{eq:prf_r2}
\end{eqnarray}
Thus, the coefficient of $e_j$ on $r_{j}^{(-j)}$ is one. 
Now, suppose that $x_j$ has a child $x_i$.  
If the coefficient of $e_j$ on $x_i$ is non-zero, $r_{j}^{(-j)}$ and $\bfx_{(-j)}$ are dependent due to D-S theorem. 
Even if it is zero, {\it i.e.}, cancelled out to be zero by special parameter values of the connection strengths,  the coefficient of $e_j$ on at least one other variable in $\bfx_{(-j)}$ is non-zero since there must be such an observed variable to cancel out the coefficient of $e_j$ on $x_i$ to be zero. 
It implies that $r_{j}^{(-j)}$ and $\bfx_{(-j)}$ are dependent due to D-S theorem.
Next, suppose that $x_j$ has a latent confounder $f_i$.
Then, in Eq.(\ref{eq:prf_r2}), the corresponding element in ${\pmb \lambda}_j$ is not zero, {\it i.e.}, the coefficient of $f_i$ on $r_j^{(-j)}$ is not zero. 
Further, $f_i$ has a non-zero coefficient on at least one variable in $\bfx_{(-j)}$ due to the definition of latent confounders.  
Therefore, $r_{j}^{(-j)}$ and $\bfx_{(-j)}$ are dependent due to D-S theorem. 

On the other hand, in Eq.(\ref{eq:prf_r}), if $\bfa_{j(-j)}^T - \bsigma^T_{(-j)j} \Sigma_{(-j)}^{-1} \A_{(-j)} \neq {\bf 0}^T$, at least one of the coefficients of the elements in $\bfe_{(-j)}$ on $r_{j}^{(-j)}$ is not zero. 
By definition, every element in $\bfe_{(-j)}$ has a non-zero coefficient on the corresponding element in $\bfx_{(-j)}$, 
Thus, $r_{j}^{(-j)}$ and $\bfx_{(-j)}$ are dependent due to D-S theorem.

 ii) The converse of contrapositive of i) is straightforward using the model definition. 
From i) and ii), the lemma is proven. 
\mbox{\hfill \bsquare}

\end{document}